\documentclass{pinepaper}
\usepackage{multirow}

\title{WorldSample: Closed-loop Real-robot RL with World Modelling}

\author[1]{Yuquan Xue}
\author[2]{Le Xu}
\author[3,1]{Zeyi Liu}
\author[4]{Zhenyu Wu}
\author[1]{Zhengyi Gu}
\author[1]{Xinyang Song}
\author[1]{Bofang Jia}
\author[1]{Ziwei Wang\textsuperscript{\textdagger}}

\affil[1]{PINE Lab, School of Electrical and Electronic Engineering, Nanyang Technological University, Singapore}
\affil[2]{Department of Electronic Engineering, Tsinghua University, Beijing, China}
\affil[3]{School of Automation, Central South University, Changsha, China}
\affil[4]{School of Automation, Beijing University of Posts and Telecommunications, Beijing, China}

\date{\small \textsuperscript{\textdagger}Corresponding author.}

\papervenue{Preprint}
\paperdate{\today}
\projectpage{https://xxreinsno.github.io/worldsample/}
\correspondence{\href{ziwei.wang@ntu.edu.sg}{ziwei.wang@ntu.edu.sg}}

\begin{document}
\makepinetitle

\begin{pineabstract}
    Reinforcement learning~(RL) can overcome the demonstration-coverage limitation of imitation learning~(IL) by allowing robots to improve through trial-and-error interaction beyond the states observed in demonstrations.
However, deploying RL on real robots remains constrained by high interaction costs, since each physical rollout is costly and reflects only one realized action-outcome path.
To address this challenge, we propose \textbf{WorldSample}, a physically grounded data augmentation framework for real-robot RL that closes a real-synthetic loop between physical rollouts, world-model generation, and policy improvement.
Grounded on real rollouts, WorldSample generates high-fidelity synthetic transitions through a post-trained world model, which greatly lowers the visual hallucination.
Specifically, rather than simply using these transitions as real-world experience, WorldSample introduces Policy-Paced Learning (PPL) to regulate the training process through sample selection and scheduling, balancing useful augmentation against value overestimation and mitigating the hallucination-induced noise.
Experiments on robot manipulation tasks involving contact-rich and precise tasks show that WorldSample improves policy success rate by 28\% while reducing training steps by 59\% compared with baselines. 
Furthermore, WorldSample improves world model visual fidelity by 19.4dB in PSNR and 0.47 in SSIM over demonstration-only post-training, validating the effectiveness of the real-synthetic loop for both policy and world model performance.
\end{pineabstract}


\section{Introduction}
\label{sec:intro}
	Imitation Learning (IL) has made substantial progress in acquiring complex manipulation skills from large-scale human demonstrations, enabling policies to generalize across diverse objects, scenes, and tasks~\cite{kim2024openvla, hung2025nora, black2026pi0, intelligence2025pi_, liu2024rdt, liu2026rdt2}.
However, IL remains limited by the coverage of its demonstration data: when a policy encounters out-of-distribution states underrepresented in the dataset, small errors can compound and lead to failure~\cite{li2024towards,saxena2025what}.
RL offers a complementary path by enabling robots to improve through trial-and-error interaction and reward-based feedback beyond demonstrated behaviors~\cite{zhang2025reinbot,ren2024diffusion,huang2025co,wagenmaker2025steering,luo2025precise}.
Yet real-world RL faces a fundamental rollout-cost bottleneck.
Physical rollouts are slow to collect, may cause hardware wear or unsafe contact, often require human reset or intervention, and each rollout reveals only one realized action-outcome path.

\begin{figure}[ht]
\centering
\includegraphics[width=1.0\textwidth]{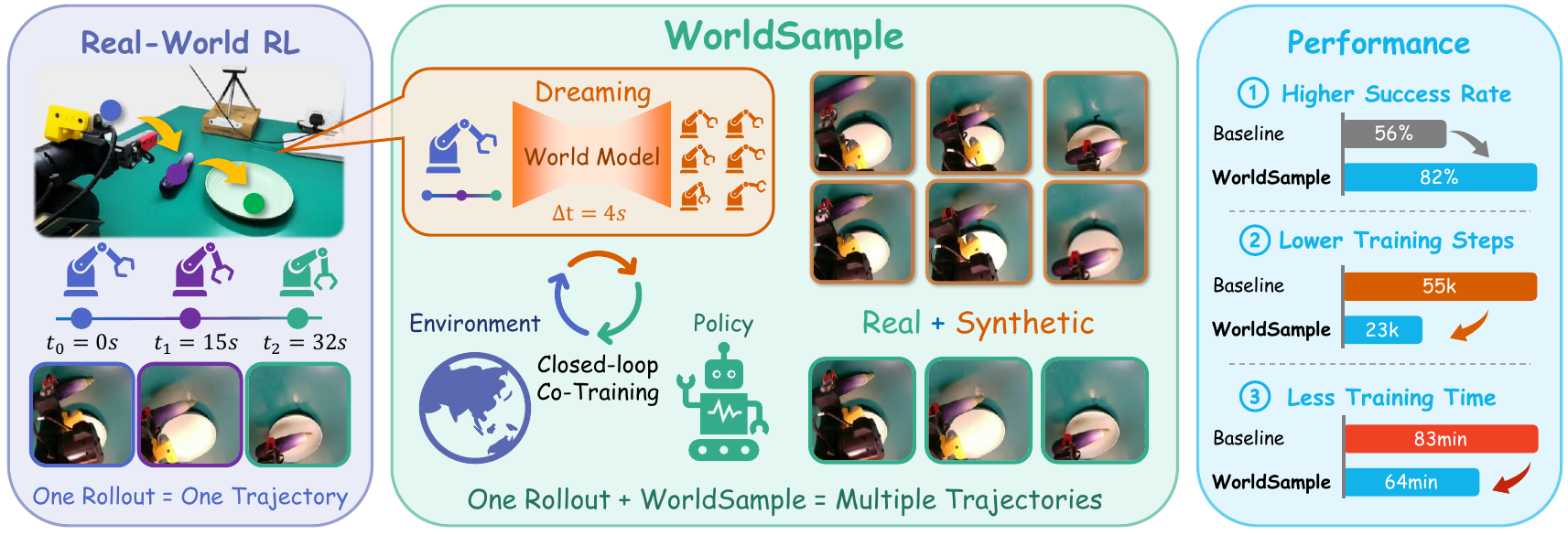}
\caption{
\textbf{WorldSample overview.}
WorldSample expands real-robot rollout into multiple physically grounded synthetic trajectories with an action-conditioned world model, forming a real-synthetic training loop that improves success rate and reduces interaction cost.
}
\label{fig:teaser}
\end{figure}

Existing methods reduce rollout cost from different points in the data pipeline.
Demonstration-based and human-in-the-loop methods use expert trajectories, corrective interventions, and prior offline data to improve initial policy quality and stabilize online training~\cite{ghasemipour2025self,liu2026rdt2,ball2023efficient,luo2025precise}.
Although effective, they still depend on costly human effort and physical rollouts, with each collected trajectory covering only a single action-outcome path.
Traditional data augmentation methods perturb image observations or action sequences to improve robustness via data variation~\cite{xue2025demogen, mandlekar2023mimicgen, xue2026resample,jin2025sime}, but these unguided methods neglect the physical reachability and reasonable dynamics.
World-model-based methods synthesize action-conditioned experience by predicting future observations, rewards, or rollout trajectories~\cite{yang2026rise, liu2026world, guo2026vlaw}.
However, when used as standalone simulators or dreaming environments, current world models can drift away from the physical data distribution, producing visual hallucinations and contact-dynamics artifacts that introduce noisy supervision for policy learning.
These limitations motivate a trajectory-level augmentation framework that expands action-conditioned experience while remaining anchored to real robot rollouts.

In this paper, we propose \textbf{WorldSample}, a physically grounded world-model augmentation framework for online real-robot learning.
As illustrated in Fig.~\ref{fig:teaser}, WorldSample expands real robot rollouts into multiple task-relevant synthetic trajectories through a real-synthetic data loop: online rollouts adapt the world model to the target task distribution, while the adapted model generates trajectories grounded in real visual observations and action sequences.
Meanwhile, to mitigate the synthetic noise introduced by world-model artifacts and prevent generated data from destabilizing learning, WorldSample introduces \textbf{Policy-Paced Learning (PPL)}, which regulates synthetic data usage through Q-selection and uncertainty-guided scheduling.
Together with asynchronous world-model generation, WorldSample expands real-world experience without replacing physical interaction or blocking real-time control.
We evaluate WorldSample on robot manipulation tasks involving contact-rich insertion and precision assembly.
Compared with the real RL baseline, WorldSample achieves faster convergence, higher success rates, and reduced physical interaction and human intervention.
In summary, our contributions are as follows:
\begin{itemize}
    \item We propose \textbf{WorldSample}, a physically grounded world-model augmentation framework to reduce the interactive cost of real-world robotic manipulation RL.
    \item We formulate a real-synthetic data loop, aligning video prediction with physical dynamics via counterfactual trajectory generation and asynchronous real-synthetic scheduling.
    \item We introduce \textbf{Policy-Paced Learning (PPL)}, which stabilizes real-synthetic training through Q-aware sample selection and uncertainty-guided data scheduling.
\end{itemize}


\section{Related Work}
\label{sec:relatedwork}
    \subsection{Real World Online RL}

Reinforcement learning (RL) has shown promising results in enabling robots to master complex manipulation behaviors through direct environmental interaction.
Recent online RL methods, such as VLA-RL~\cite{lu2025vla}, SimpleVLA-RL~\cite{li2025simplevla}, and RLinf-VLA~\cite{zang2025rlinfvla}, leverage policy optimization to continuously improve across multiple tasks.
However, learning these behaviors entirely from scratch requires a massive number of physical rollouts, making them highly sample-inefficient for real-world deployment.
To mitigate this, SAC-based methods like HIL-SERL~\cite{luo2025precise} and ConRFT~\cite{chen2025conrft} integrate offline datasets and human interventions with online rollouts to accelerate convergence.
Despite these improvements, they still heavily depend on expensive physical interactions and costly human resources for continuous data collection and timely intervention.
In summary, the fundamental bottleneck for deploying RL in real-world robotic manipulation remains the prohibitively high cost of physical trial-and-error training.

\subsection{World Models Based Interaction}

Recent action-conditioned world models have enabled closed-loop interaction between robot policies and learned environments.
Early works such as VLA-RFT~\cite{li2025vla}, ProphRL~\cite{zhang2025reinforcing}, WoVR~\cite{jiang2026wovr} and World-Env~\cite{xiao2025world} mainly study policy optimization inside simulated or generated environments, using world-model rollouts with objectives such as GRPO~\cite{shao2024deepseekmath} or LOOP-style optimization~\cite{chen2025reinforcement}.
These methods show the potential of coupling world models with policy improvement, but transferring such pipelines to physical robots remains challenging since synthetic environments are hard to model real dynamics.
More recent real-robot-oriented methods, including World-VLA-Loop~\cite{liu2026world}, RISE~\cite{yang2026rise}, VLAW~\cite{guo2026vlaw}, and WMPO~\cite{WMPO2025}, use world models as simulators or synthetic data generators for policy improvement, either through IL or RL.
However, when policy learning relies heavily on imagined rollouts, visual hallucination and action-following errors can create a gap between generated scenarios and real interactions.


\section{WorldSample}
\label{sec:method}
    \subsection{WorldSample Pipeline}
\label{sec:method_overview}

\begin{figure}[t]
\centering
\includegraphics[width=1.0\textwidth]{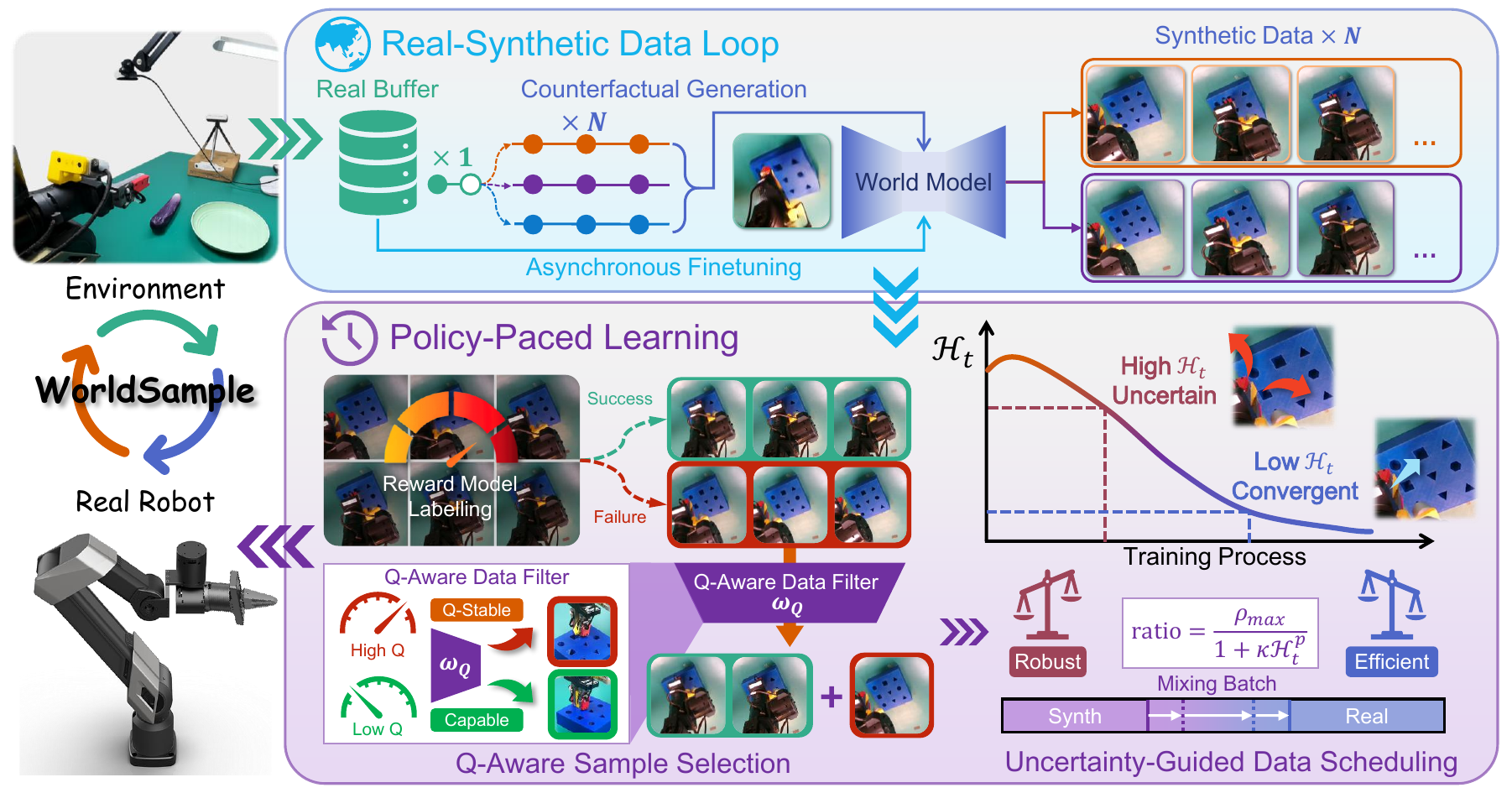}
\caption{
\textbf{WorldSample pipeline.}
WorldSample closes a real-synthetic data loop between physical rollouts, world-model generation, and policy improvement, while Policy-Paced Learning (PPL) regulates the composition and usage of synthetic data during training.
}
\label{fig:pipeline}
\end{figure}

We propose \textbf{WorldSample}, a physically grounded, closed-loop data augmentation framework for online real-world RL.
As shown in Fig.~\ref{fig:pipeline}, WorldSample consists of two tightly coupled components: \textbf{real-synthetic data loop} and \textbf{Policy-Paced Learning (PPL)}.

The real-synthetic data loop starts from physical robot rollouts collected during online interaction.
These rollouts serve two roles simultaneously: they provide training data for policy learning, and they ground synthetic data generation through a action-conditioned world model.
Conditioned on real observations and counterfactual sampled trajectories, the world model generates synthetic transitions labelled by a reward model.
As new rollouts accumulate, they are further used to post-train the world model, closing the loop between physical interaction and synthetic generation.

PPL regulates how the generated data is incorporated into policy learning.
It includes Q-aware sample selection, which controls the composition of generated transitions, and uncertainty-guided scheduling, which determines how much synthetic data should be introduced during RL training.
Together, these components allow WorldSample to enlarge the effective training distribution while suppressing value overestimation and hallucination-induced noise.

Overall, WorldSample uses real interaction to anchor world-model generation and uses PPL to safely exploit synthetic data for online robot learning.
This design improves RL training without replacing real interaction or violating real-time deployment constraints.

\subsection{Real-Synthetic Data Loop}
\label{sec:method_generation}

\paragraph{Counterfactual Trajectory Generation.}
WorldSample generates synthetic transitions by sampling counterfactual trajectory segments around physically executed rollouts. 
Rather than drawing actions from a random action prior, we sample trajectory segments from the real rollout distribution with local perturbations.
Given a real rollout segment $\tau=(o_0,A_{0:T-1},O_{1:T})\sim\mathcal{D}_{\mathrm{real}}$, we have:
\begin{equation}
\tilde{A}_{0:T-1}\sim q_{\sigma,s}(\tilde{A}_{0:T-1}\mid A_{0:T-1})
\label{eq:counterfactual_action_sampling}
\end{equation}
where $q_{\sigma,s}$ denotes a scale-randomized local perturbation distribution centered on real action sequence. 
This counterfactual sampling preserves the temporal structure and action statistics of robot policy while expanding data coverage, making generated trajectories more physically feasible than unconstrained rollouts.
The world model further predicts the visual future, and the reward model assigns task feedback:
\begin{equation}
\tilde{O}_{1:T}\sim p_{\phi}(\cdot\mid o_0,\tilde{A}_{0:T-1}),
\qquad
\tilde{r}_{0:T-1}=\hat{R}_{\psi}(\tilde{O}_{1:T}).
\label{eq:world_model_counterfactual_rollout}
\end{equation}
The resulting tuple $(o_0,\tilde{A}_{0:T-1},\tilde{O}_{1:T},\tilde{r}_{0:T-1})$ defines a labeled synthetic trajectory. 
Since both the visual and action conditions are derived from physical rollouts, the generated data remains grounded to the physical dynamics and real geometry while providing trajectory-level diversity.

\paragraph{Asynchronous Scheduling.}
WorldSample executes counterfactual generation asynchronously to preserve real-time robot control. 
The robot learning stream continues collecting physical rollouts and updating the policy, while the world-model generates synthetic trajectories in parallel. 
This decoupling prevents slow video generation from blocking policy execution and allows multiple world-model workers to increase data throughput during online training.
Finally, the loop is closed by asynchronously adapting the world model with online rollout data.
As new rollouts accumulate, the world-model stream incorporates them into post-training, aligning generation with high-diversity data coverage.
In this way, real interaction continually improves the world model, and the adapted world model provides a higher-quality synthetic experience for subsequent policy learning.

\subsection{Policy-Paced Learning}
\label{sec:method_ppl}

While synthetic data expands the physical experience distribution, naively injecting generated transitions into robot learning can destabilize value learning.
Let $\mathcal{D}_{\mathrm{real}}$ denote the physical replay buffer containing online rollouts and demonstrations, and $\mathcal{D}_{\mathrm{syn}}$ denote generated transitions.
Since a post-trained world model remains an imperfect approximation of real dynamics, each synthetic transition carries residual error $\epsilon_{\mathrm{syn}}(s,a)$.
When the critic trains on such transitions, local errors can compound through Bellman backups:
\begin{equation}
\Delta Q(s,a)
\;\approx\;
\mathbb{E}
\left[
\sum_{k=0}^{\infty}
\gamma^k
\epsilon_{\mathrm{syn}}(s_k,a_k)
\right],
\label{eq:critic_bias}
\end{equation}
Thus, naively injecting synthetic transitions can induce severe Q-value overestimation and extrapolation errors.

To stabilize RL training, we propose \textbf{Policy-Paced Learning (PPL)}, which separates synthetic data usage into sample-level value balancing and policy-level scheduling:
\begin{equation}
\mathcal{L}_{Q}^{\mathrm{PPL}}(\theta)
=
\mathbb{E}_{\tau\sim \mathcal{D}_{\mathrm{real}}}
[
\ell_Q(\tau;\theta)
]
+
\rho(H_t)
\,
\mathbb{E}_{\tilde{\tau}\sim \mathcal{D}_{\mathrm{syn}}}
[
w_Q(\tilde{\tau})\ell_Q(\tilde{\tau};\theta)
].
\label{eq:ppl_critic_objective}
\end{equation}
where $\ell_Q$ is the TD loss, $w_Q$ controls the composition of synthetic supervision, and $\rho(H_t)$ controls its training-stage-dependent influence.
As the actor is optimized through the learned critic, stabilizing the synthetic critic target also stabilizes the downstream actor update.

\paragraph{Q-Aware Sample Selection.}

Q-aware selection controls the value signal induced by generated data.
Positive synthetic trajectories expand rare successful outcomes and accelerate value propagation, while negative trajectories constrain visually plausible but unsuccessful behaviors and suppress optimistic extrapolation.
Each generated trajectory is labeled by an independent reward model, yielding positive and negative sets $\mathcal{D}_{\mathrm{syn}}^{+}$ and $\mathcal{D}_{\mathrm{syn}}^{-}$.
PPL selects synthetic data to preserve a value-balanced contrast between positive and negative samples:
\begin{equation}
\left|
\left(
\mathbb{E}_{\tilde{\tau}\sim\mathcal{D}_{\mathrm{syn}}^{+}}
Q_{\theta}(\tilde{s},\tilde{a})
+
\mathbb{E}_{\tilde{\tau}\sim\mathcal{D}_{\mathrm{syn}}^{-}}
Q_{\theta}(\tilde{s},\tilde{a})
\right)
-
\mathbb{E}_{\tau\sim \mathcal{D}_{\mathrm{real}}}
[
Q_{\theta}(s,a)
]
\right|
\leq \delta_Q.
\label{eq:q_aware_selection}
\end{equation}
where $\Delta Q_{\mathrm{real}}$ denotes the corresponding value contrast estimated from physically grounded experience.
This prevents synthetic data from shifting the critic toward uniformly optimistic or overly conservative estimates.

\paragraph{Uncertainty-Guided Data Scheduling.}
Even with value-balanced selection, synthetic transitions remain noisier than physical rollouts.
PPL therefore schedules its influence according to policy uncertainty, measured by actor entropy on real states:
\begin{equation}
H_t
=
\mathbb{E}_{s\sim \mathcal{D}_{\mathrm{real}}}
\left[
\mathcal{H}(\pi_t(\cdot|s))
\right]
=
-
\mathbb{E}_{s\sim \mathcal{D}_{\mathrm{real}}, a\sim \pi_t}
\left[
\log \pi_t(a|s)
\right].
\label{eq:actor_entropy}
\end{equation}
High $H_t$ indicates that the policy remains uncertain on real states, making actor-critic learning more sensitive to biased synthetic targets.
Since the synthetic term in Eq.~\ref{eq:ppl_critic_objective} directly perturbs critic learning and the actor is optimized through the learned critic, the effective value bias induced by synthetic supervision can be summarized as
\begin{equation}
\Delta Q_{\mathrm{syn}}(t)
\propto
\rho(H_t)
\cdot
H_t
\cdot
\mathbb{E}_{\tilde{\tau}\sim \mathcal{D}_{\mathrm{syn}}}
\left[
w_Q(\tilde{\tau})\epsilon_{\mathrm{syn}}(\tilde{\tau})
\right],
\label{eq:synthetic_bias_entropy}
\end{equation}
This relation motivates an entropy-paced schedule that suppresses synthetic influence under high policy uncertainty:
\begin{equation}
\rho(H_t)
=
\rho_{\max}
\cdot
\frac{1}{1+\kappa H_t^p},
\qquad
\kappa>0,\; p\geq 1.
\label{eq:entropy_paced_ratio}
\end{equation}
The exponent $p$ controls how conservatively synthetic data is introduced, while $\rho_{\max}$ prevents generated data from dominating physical experience. Equivalently, PPL controls the effective synthetic bias as
\begin{equation}
\mathrm{Bias}(t)
\propto
\frac{\rho_{\max} H_t}
{1+\kappa H_t^p}
\cdot
\mathbb{E}
\left[
\|\epsilon_{\mathrm{dyn}}\|
\right]
\;\longrightarrow\; 0
\quad
\text{as}
\quad
H_t\rightarrow 0 .
\label{eq:effective_bias}
\end{equation}
Thus, synthetic transitions is used conservatively when the policy is uncertain and is gradually introduced as the policy becomes more concentrated on real robot states. 
Together, Q-aware selection and uncertainty-guided scheduling separate \textit{which} synthetic trajectories are admitted from \textit{when} they are trusted, improving sample efficiency while mitigating Q-value overestimation and synthetic noise.


\section{Experimental Results}
\label{sec:result}
    In this section, we conduct extensive experiments on complex and real-world tasks, including precise manipulation and contact-rich interaction, to evaluate WorldSample.
Our experiments are designed to answer the following three questions:
\begin{itemize}
    \item Does WorldSample improve online real-robot RL efficiency?
    \item Does WorldSample improve the video fidelity and physical grounding of the world model?
    \item Is Policy-Paced Learning necessary for stable and effective use of synthetic data?
\end{itemize}

\subsection{Experimental Settings}

\paragraph{Experimental Setups.}
We conduct experiments on a Galaxea A1X robotic arm equipped with a binary gripper. 
The robot observes the scene through two Intel RealSense D435i cameras: a side-view camera for third-person perception and a wrist-mounted camera for egocentric perception. 
We evaluate WorldSample on five categories of real-world manipulation tasks with different levels of contact complexity, precision requirement, and action-space difficulty, including \textbf{Pushing}, \textbf{Insertion}, \textbf{Sorting}, \textbf{Pick \& Place}, and \textbf{Assemble}.
Further task details are provided in Appendix~X.

\paragraph{Baselines and Training Details.}
We compare WorldSample with a human-in-the-loop real-robot RL baseline following HIL-SERL~\cite{luo2025precise}, and state-of-the-art world-model co-training methods VLAW~\cite{guo2026vlaw} and WMPO~\cite{WMPO2025}. 
All methods use the same robot platform, observation setup, action space, reward definition, intervention protocol, and 20 human demonstrations for initialization, while VLAW leverages an additional 50 synthetic demonstrations.
The policy is trained with a total batch size of 256.
For the world-model stream, we use the action-conditioned Cosmos-Predict2.5 model as the base video world model. 
The model is first adapted with task-specific human demonstrations and then further refined with online RL rollouts.
All model-based methods use the same adapted world model for fair comparison.
Implementation details are provided in the Appendix.

\subsection{Experimental Results}

\paragraph{Real-World Experiments.}
We first evaluate whether WorldSample improves online real-robot RL efficiency and policy capability.
All methods use the same demonstration budget, intervention protocol, and real-robot setup.
For each task, the baselines train till its convergence and define the reference real-robot training budget, while WorldSample's training is stopped once the policy reaches convergence or when its wall-clock training time matches the corresponding baselines.

\begin{table}[t]
\centering
\small
\caption{
Real-world Experimental results.
}
\label{tab:main_results}
\resizebox{0.95\textwidth}{!}{
\begin{tabular}{llcccccc}
\toprule
Method & Metric & Pushing & Insertion & Sorting & Pick\&Place & Assemble & Avg. \\
\midrule
\textbf{VLAW} & Success Rate & 86\% & 47\% & 78\% & 76\% & 32\% & 64\% \\
\midrule
\multirow{2}{*}{\textbf{WMPO}}
& Success Rate & 90\% & 82\% & 72\% & 78\% & 23\% & 69\% \\
& Training Time & 140min & 484min & 468min & 801min & 840min & 547min \\
\midrule
\multirow{3}{*}{\textbf{HIL-SERL}}
& Success Rate       & 84\% & 63\% & 66\% & 55\% & 10\% & 56\% \\
& Training Steps     & 15K  & 40K  & 52K  & 78K  & 96K  & 56K \\
& Training Time      & 30min  & 60min  & 75min  & 110min & 140min & 83min \\
\midrule
\multirow{3}{*}{\textbf{WorldSample}}
& Success Rate       & 95\% & 95\% & 95\% & 84\% & 42\% & 82\% \\
& Training Steps     & 8K   & 10K  & 20K  & 36K  & 40K  & 23K \\
& Training Time      & 24min  & 30min  & 50min  & 100min & 120min & 64min \\
\bottomrule
\end{tabular}
}
\end{table}

\begin{figure}[t]
\centering
\includegraphics[width=1.0\textwidth]{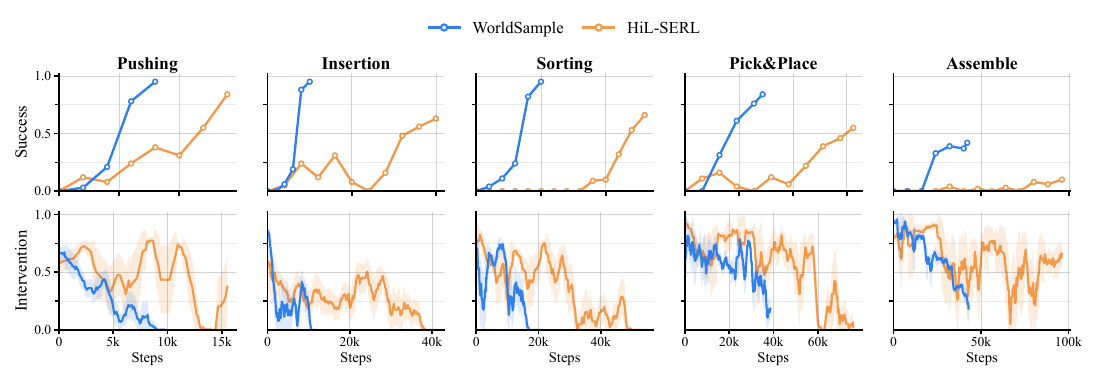}
\caption{Success Rate and Intervention Rate for each task.}
\label{fig:main_results}
\end{figure}

Table~\ref{tab:main_results} summarizes the experimental results.
Compared with HIL-SERL, WorldSample significantly improves the success rate on all five tasks, increasing the average success from 56\% to 82\%.
It also reduces the average number of training steps from 56K to 23K, and training time from 83 minutes to 64 minutes, corresponding to a 59\% reduction in computational cost and a 23\% reduction in physical interactions.
The improvement is especially clear on Insertion and Sorting, where WorldSample reaches 95\% success with 50\% and 33\% fewer real interaction steps, respectively.
Compared with model-based baselines, WorldSample achieves substantially higher success on all tasks, especially on Insertion and Assemble, where the performance of model-based methods is constrained by the data quality of generated trajectories, while the performance gap on Pushing and Pick\&Place is smaller since the relatively easier dynamics and visual distribution of these tasks are better captured by the world model.

Fig.~\ref{fig:main_results} shows the corresponding online training curves.
WorldSample generally reaches higher success earlier in training, indicating that the generated transitions provide useful learning signals under limited real-world interaction.
On Pushing, Insertion, and Sorting, WorldSample significantly improves both success and training time trends.
On Pick\&Place and Assemble, WorldSample achieves substantially higher success, while the time cost remains comparable to HIL-SERL due to the increased difficulty of long-horizon contact-rich execution.

Overall, these results show that WorldSample improves online real-robot RL primarily by extracting more learning signals from each physical rollout.
Although training time reduction is task-dependent on harder long-horizon tasks, WorldSample consistently achieves higher final success under a smaller real-world training cost.

\paragraph{World Model Adaptation.}

\begin{figure}[t]
\centering
\includegraphics[width=1.0\textwidth]{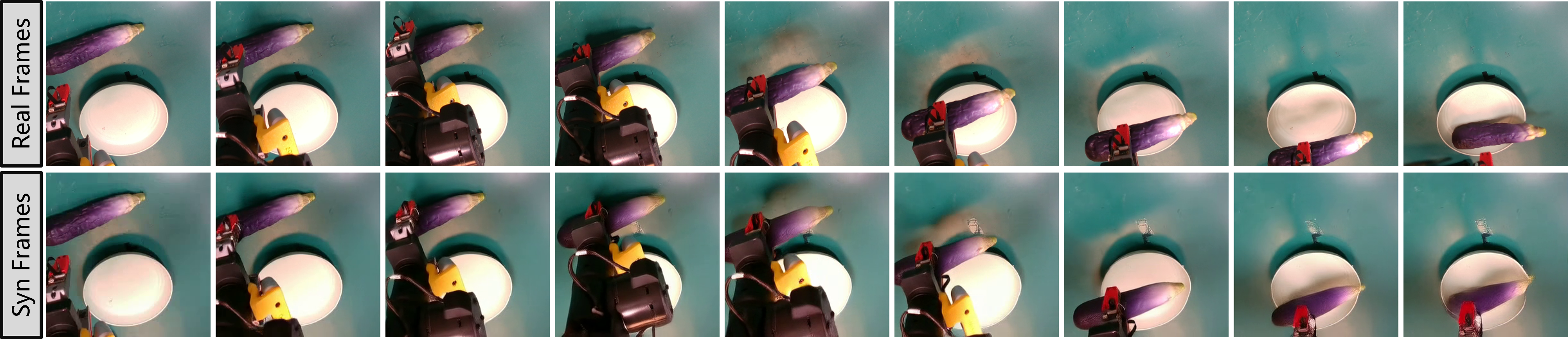}
\caption{Comparison between real and generated video frames.}
\label{fig:visualization}
\end{figure}

Table~\ref{tab:wm_ablation}.a evaluates the world model under different adaptation data. 
The pretrained model performs poorly in the target robot domain, indicating a large gap between generic video priors and the task-specific visual-action distribution. 
Finetuning only on human demonstrations provides limited improvement, since human demonstrations only cover a narrow set of action space. 
In contrast, online rollout adaptation substantially improves prediction quality, increasing PSNR from 8.501 to 28.428 and SSIM from 0.334 to 0.906, while reducing LPIPS from 0.7432 to 0.0343.
This suggests that online RL data provides more diverse state-action outcomes, including failures and recovery behaviors, which helps the world model generalize beyond the demonstration distribution. 
Combining human demonstrations with online rollouts further improves PSNR and SSIM, and the dual-view model achieves the best structural fidelity with 29.887 PSNR and 0.925 SSIM, since the concatenated view supports more supervision.
As shown in Fig.~\ref{fig:visualization}, the rollout-adapted model can generate realistic videos with accurate object shapes and positions.
These results show that online rollout adaptation is critical for making the world model a physically grounded engine for WorldSample.

\begin{table}[t]
\centering
\scriptsize
\caption{
\textbf{World-model quality and ablation.}
}
\label{tab:wm_ablation}
\begin{minipage}[t]{0.45\textwidth}
\centering
\caption*{(a) World-model prediction quality}
\resizebox{\linewidth}{!}{
\begin{tabular}{lccc}
\toprule
Variant & PSNR & SSIM  & LPIPS  \\
\midrule
Pretrained & 8.50 & 0.334 & 0.743 \\
Demo-only & 10.27 & 0.428 & 0.809 \\
Rollout-only & 28.43 & 0.906 & \textbf{0.034} \\
Demo+Rollout & 29.66 & 0.910 & 0.035 \\
Dual-view & \textbf{29.89} & \textbf{0.925} & 0.035 \\
\bottomrule
\end{tabular}
}
\end{minipage}
\hfill
\begin{minipage}[t]{0.534\textwidth}
\centering
\caption*{(b) PPL ablation on insertion task}
\resizebox{\linewidth}{!}{
\begin{tabular}{lcccc}
\toprule
Setting & Succ. & Intv.  & Steps  & Time  \\
\midrule
\textbf{WorldSample} & \textbf{95\%} & 24\% & \textbf{10K} & \textbf{30min} \\
w/o Scheduling & 61\% & 43\% & 18K & 48min \\
w/o Q-Selection & 76\% & 42\% & 12K & 36min \\
w/o Full-PPL & 86\% & 23\% & 32K & 69min \\
HIL-SERL & 63\% & 27\% & 40K & 60min \\
\bottomrule
\end{tabular}
}
\end{minipage}
\end{table}

\subsection{Ablation Studies}

\begin{figure}[t]
\centering
\includegraphics[width=1.0\textwidth]{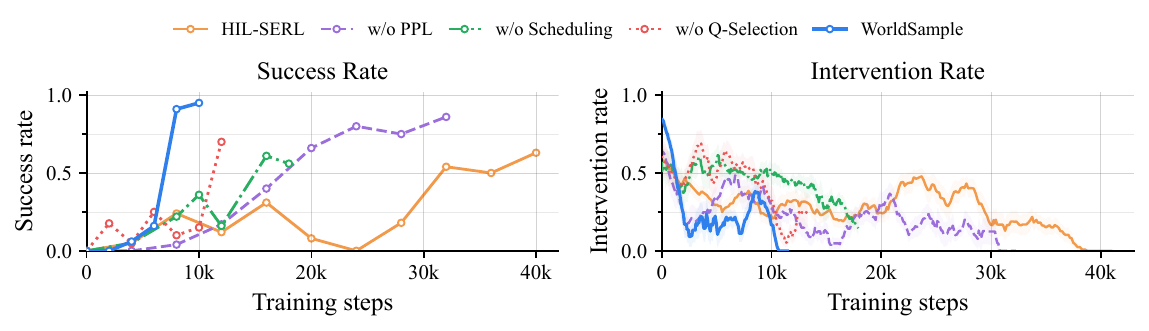}
\caption{
Success Rate and Intervention Rate on each ablation setting.
}
\label{fig:ablation}
\end{figure}

We further evaluate whether Policy-Paced Learning is necessary for stable use of synthetic trajectories.
Table~\ref{tab:wm_ablation}.b reports ablations on the insertion task, where precise contact and complex dynamics make it especially sensitive to the data quality and scheduling.
Naive strategy(w/o PPL) may eventually improve success to 86\%, but requires 32K steps and 69 minutes, suggesting that synthetic data is useful but inefficient without policy-paced control.
Removing uncertainty-guided scheduling reduces success from 95\% to 61\%, showing that synthetic data should not be naively introduced throughout online training, as visual hallucination can lead to noisy policy capability.
Removing Q-aware sample selection also degrades performance to 76\% success and increases intervention rate to 42\%, indicating that the composition of generated outcomes is important for preventing unstable value learning.

Fig.~\ref{fig:ablation} further illustrates the mechanism behind these results.
The full method and baseline share similar intervention trends, while the full method achieves higher success with fewer training steps, indicating that the generated trajectories provide useful learning signals under limited real-world interaction.
Removing scheduling leads to a significant drop in success, indicating that introducing synthetic data without scheduling can lead to unstable training and suboptimal policy capability.
Removing Q-aware sample selection also degrades performance, showing that the composition of generated trajectories is important for preventing unstable value learning.
Together, these results validate the two roles of PPL introduced in Sec .~\ref {sec:method_ppl}: Q-aware sample selection controls the value composition of generated trajectories, while uncertainty-guided scheduling controls how much synthetic data is used.


\section{Conclusion}
\label{sec:conclusion}
    We presented \textbf{WorldSample}, a physically grounded data augmentation framework for online real-robot RL. 
WorldSample expands physical rollout to synthetic transitions through world models, while keeping real interaction as the grounding source. 
PPL further regulates generated data through Q-aware selection and uncertainty-guided scheduling, enabling useful augmentation while mitigating value overestimation and hallucination-induced noise. 
Experiments on contact-rich and precision manipulation tasks show improved success, faster convergence, and reduced real interaction cost over baselines.

\section{Limitaions}
\label{sec:limitations}
    WorldSample currently focuses on augmenting real-robot learning within a single task and a relatively fixed scene distribution.
Although online rollout adaptation improves the world model within this setting, extending this framework to instance-, task-, or scene-level generalization remains an important direction for future work.

\clearpage
\bibliographystyle{plain}
\bibliography{ref}


\clearpage
\appendix
    \section{Experiment Details}

\subsection{Real-World Experimental Setup}
\label{app:setup}

\begin{figure}[ht]
\centering
\includegraphics[width=1.0\textwidth]{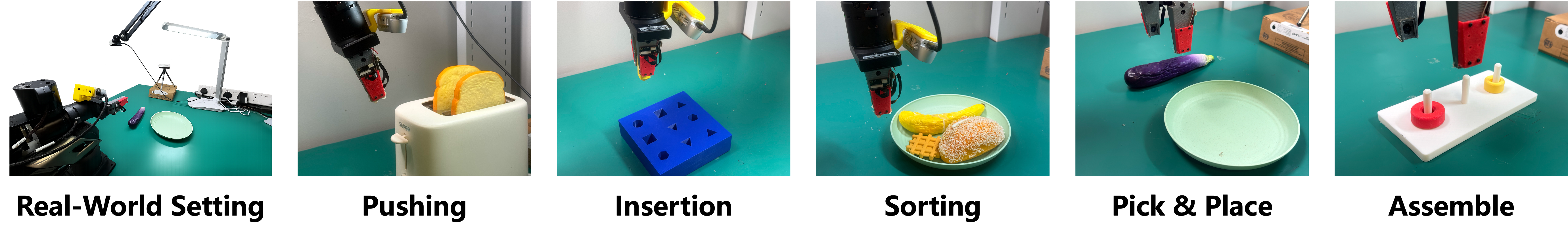}
\caption{
\textbf{Real-world task settings.}
We evaluate WorldSample on diverse manipulation tasks covering contact-rich insertion, object pushing, sorting, pick \& place, and precision assembly. These tasks require different combinations of visual grounding, contact reasoning, precise motion, and recovery from imperfect policy executions.
}
\label{fig:app_setting}
\vspace{-0.3cm}
\end{figure}

In this paper, we evaluate WorldSample and all baselines on a suite of real-world manipulation tasks, as illustrated in Fig.~\ref{fig:app_setting}. The experimental platform consists of a 6-DoF Galaxea A1X robot arm with a binary gripper, and two Intel RealSense D435i cameras providing wrist and third-person views. The tasks are designed to cover a range of manipulation challenges, including contact-rich interactions, precision alignment, visual discrimination, and long-horizon manipulation. The specific task categories include:
\begin{itemize}
    \item \textbf{Pushing.} Press the toaster button with bread in the toaster.
    \item \textbf{Insertion.} Insert a shaped block into the corresponding hole.
    \item \textbf{Sorting.} Pick up the target object from a plate containing distractor objects.
    \item \textbf{Pick \& Place.} Pick the target object and place it into a container.
    \item \textbf{Assemble.} Remove a ring from a Tower-of-Hanoi setup and assemble it onto another pillar.
\end{itemize}

Each task starts from 20 human demonstrations and proceeds with online robot interaction under human intervention when necessary, or real-world evaluation process.
All tasks share a common evaluation protocol, intervention strategy, and performance metrics to ensure a consistent comparison.

We report success rate, convergence training time of all RL methods, and online training steps of real-robot interaction methods.
A rollout is counted as successful if the task-specific goal condition is reached within the episode horizon; intervention rate measures the fraction of episode execution requiring human correction.
\begin{table}[h]
\centering
\small
\caption{
Summary of real-world task categories.
}
\label{tab:app_task_summary}
\begin{tabular}{lcc}
\toprule
Task Category & Main Challenge & Evaluation Signal \\
\midrule
Pushing & Contact-rich object displacement & Object reaches target region \\
Insertion & Precision alignment and contact & Object inserted into target slot/hole \\
Sorting & Visual discrimination and placement & Correct object-category placement \\
Pick-and-Place & Grasping and target placement & Object placed at target location \\
Assembly & Long-horizon precision manipulation & Assembly goal completed \\
\bottomrule
\end{tabular}
\vspace{-0.2cm}
\end{table}

\subsection{Baseline Methods}

We compare WorldSample with three representative baselines under the same robot platform, observation space, action space, and evaluation protocol. 
For fairness, all methods use the same HIL-SERL actor architecture and are evaluated on the same real-world task suite.
The main difference lies in how each method uses real and synthetic data for policy improvement.

\paragraph{HIL-SERL.}
HIL-SERL is our real-robot RL baseline. 
It builds on RLPD-style off-policy actor-critic learning, using a mixture of prior demonstrations, online rollouts, and human interventions for sample-efficient real-world training. 
Each task is initialized with 20 human demonstrations, after which the policy improves through online robot interaction with human intervention when necessary. 
We keep the original HIL-SERL training pipeline unchanged and report its success rate, intervention rate, physical training steps, and wall-clock training time.

\paragraph{VLAW.}
VLAW studies iterative co-improvement between a VLA policy and an action-conditioned video world model, where real rollout data improves the world model and generated synthetic rollouts further improve the policy. 
In our comparison, we adapt VLAW as an imitation-learning baseline under the same HIL-SERL actor architecture. 
Since VLAW uses synthetic successful rollouts for policy improvement, we train the actor with behavior cloning on 20 real demonstrations and 50 generated successful trajectories, which approximately matches the amount of online data used by the real-robot methods. 
All VLAW policies are trained for 20K behavior-cloning steps before real-world evaluation.

\paragraph{WMPO.}
WMPO performs world-model-based policy optimization by using generated rollouts for policy improvement without requiring real- world interaction. 
We implement WMPO on the same HIL-SERL actor and use the same world model with our method to provide synthetic interaction data for policy optimization. 
Following the task difficulty and online training steps, we train WMPO for 10K policy optimization steps on the three easier tasks and 20K steps on the two harder tasks, and stop training once the policy converges in the world model.
This baseline evaluates whether policy optimization driven primarily by world-model rollouts can transfer to our real-robot manipulation setting.

\section{Implementation Details}
\label{app:implementation}

\subsection{WorldSample Framework}
WorldSample is built on the HIL-SERL real-robot learning framework. The real replay buffer contains online environment rollouts, while the demonstration buffer stores expert demonstrations and human-guided corrective actions. Unless otherwise specified, the underlying SAC/RLPD learner, actor-critic architecture, and optimization hyperparameters follow the baseline implementation; WorldSample only augments the data stream through synthetic generation and Policy-Paced Learning.
To organize the generated data, WorldSample reproduces the synthetic buffer structure following the real replay buffer.

\subsection{Counterfactual Trajectory Generation}
Each action is represented as a 7-dimensional delta end-effector command,
\[
a_t = [dx,dy,dz,dr_x,dr_y,dr_z,d_{\mathrm{gripper}}].
\]
Given a real action sequence $A_{0:T-1}$, WorldSample samples local counterfactual action sequences around the executed trajectory.
The perturbation preserves the temporal structure of the real rollout while introducing bounded local variations.
Inactive action dimensions are kept fixed at zero, and all actions are projected back to the valid action range.
Thus, the counterfactual generation process can be formalized as:
\[
a't = \Pi{\mathcal{A}}\left(a_t \odot \xi + \epsilon_t \right),
\quad
\xi_j \sim \mathcal{U}(1-s,1+s),
\quad
\epsilon_{t,j}\sim \mathcal{N}(0,\sigma^2).
\]
where $\xi$ applies dimension-wise scale jitter consistently across the trajectory segment, $\epsilon_t$ provides local perturbation, and $\Pi_{\mathcal{A}}$ projects the perturbed action back to the action space. Inactive action dimensions are kept fixed at zero.
We set $s=0.20$ and $\sigma=0.05$ in all experiments. This trajectory-level perturbation produces counterfactual actions that are diverse but still physically close to real robot behavior.

\subsection{World Model Adaptation}
\begin{figure}[h]
\centering
\includegraphics[width=1.0\textwidth]{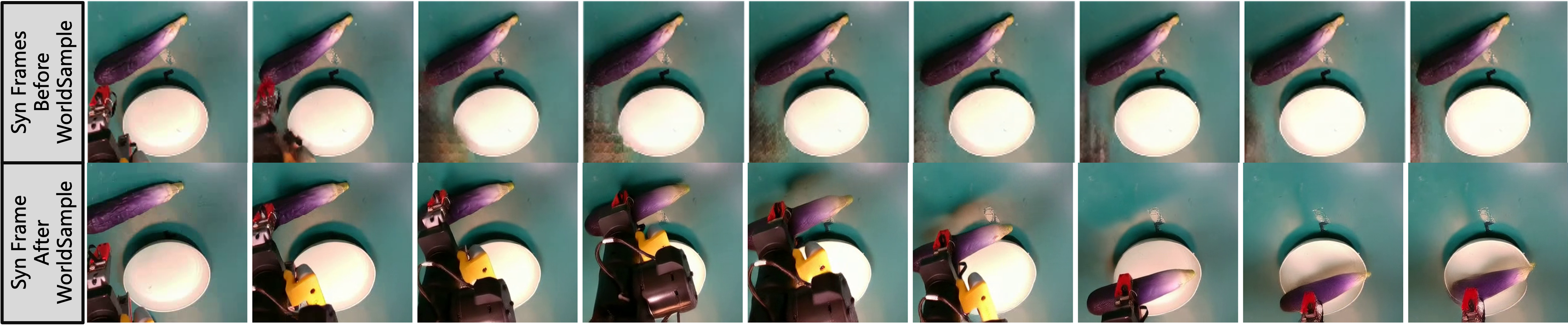}
\caption{
\textbf{World model generation quality comparison.}
Fine-tuning the world model with both demonstrations and online rollouts improves video generation quality compared with demonstration-only fine-tuning.
}
\label{fig:wm_comparison}
\vspace{-0.3cm}
\end{figure}

We initialize the video world model from action-conditioned Cosmos-Predict2.5, an open-source model pretrained on large-scale video data.
The model is then adapted with task-specific robot trajectories.
Demonstration rollouts provide initial task grounding, while online rollouts further align the model with the policy-induced state distribution, object appearance, and contact dynamics.
During online training, the world-model stream asynchronously generates candidate trajectories conditioned on real visual observations and counterfactual action sequences. To evaluate generation fidelity, we replay held-out real rollout segments with their corresponding action sequences and compute PSNR, SSIM, and LPIPS between predicted and real videos. Fig.~\ref{fig:wm_comparison} compares world models adapted with demonstrations only and with mixed demonstration-online data, showing that online rollout adaptation substantially improves video fidelity.

\subsection{Policy-Paced Learning}

\paragraph{Q-Aware Sample Selection}
Q-aware sample selection uses reward model labels to balance successful and failed generated trajectories, reducing systematic value shift.
\[
\hat{y}(\tilde{\tau}) = \mathbb{I}\left[\hat{R}_{\psi}(\tilde{\tau}) \geq \delta_R\right]
\]
where $\hat{y}=1$ denotes a successful synthetic trajectory and $\hat{y}=0$ denotes a failed one.
We store successful and failed candidate trajectories in separate queues before entering the synthetic buffer, and the data composition is controlled by Q-value signals from the critic network.
This allows PPL to balance positive samples, which propagate sparse successful outcomes, and negative samples, which constrain overly optimistic Q-values on unsuccessful but visually plausible trajectories.
In practice, Q-aware selection controls the admitted composition of these buffers to avoid a systematic value shift caused by synthetic data.

\paragraph{Uncertainty-Guided Scheduling}
Uncertainty-guided scheduling controls the mixed batch size according to actor entropy on real robot states.
In implementation, we use a linear approximation functioning as $\rho(\cdot)$ and cap the maximum synthetic ratio $\rho_{\max}=0.30$ to prevent generated data from dominating physical experience.

\section{Policy-Paced Learning Analysis}
\label{app:ppl_analysis}

\subsection{Uncertainty-Guided Scheduling}
\begin{figure}[h]
\centering
\includegraphics[width=0.85\textwidth]{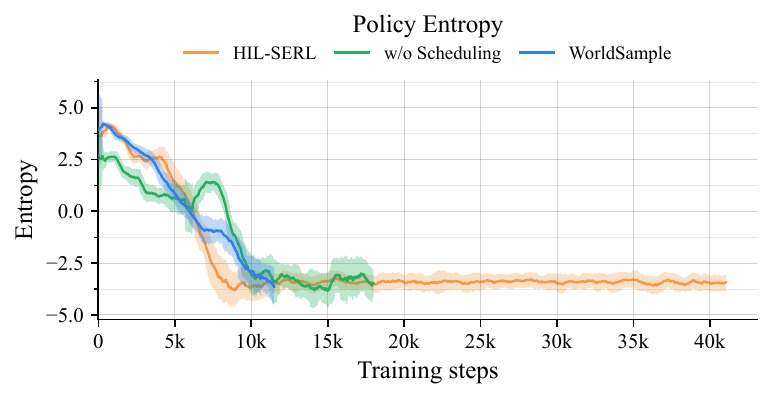}
\caption{
\textbf{Entropy-guided scheduling analysis.}
Actor entropy tracks the policy's uncertainty during training and determines how strongly synthetic trajectories are introduced. This mechanism prevents early-stage synthetic bias while allowing generated data to improve sample efficiency after the policy becomes more stable.
}
\label{fig:app_entropy}
\vspace{-0.3cm}
\end{figure}

Fig.~\ref{fig:app_entropy} provides additional analysis of the uncertainty-guided scheduling mechanism. Actor entropy serves as a policy-uncertainty signal over physically grounded states. WorldSample uses synthetic data conservatively when the policy is uncertain, and gradually increases its contribution as the policy becomes more concentrated during training.

\subsection{Q-Aware Sample Selection}
\begin{figure}[h]
\centering
\includegraphics[width=0.85\textwidth]{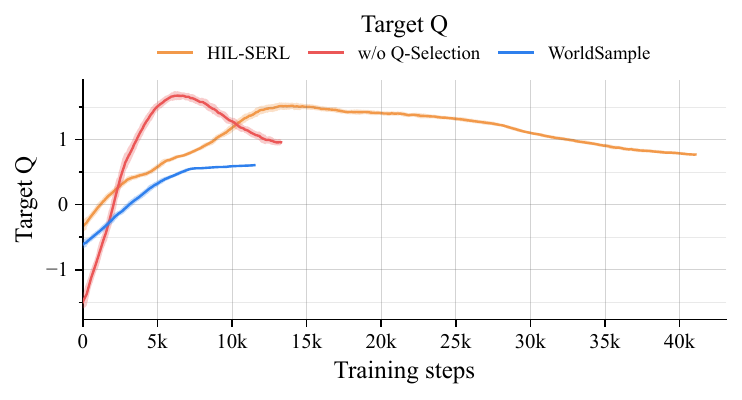}
\caption{
\textbf{Q-aware sample selection analysis.}
Balancing generated successes and failures stabilizes value learning and mitigates Q-value overestimation. Compared with naive synthetic data usage, PPL maintains more stable value estimates during online training.
}
\label{fig:app_target_q}
\vspace{-0.3cm}
\end{figure}

Fig.~\ref{fig:app_target_q} shows additional analysis of Q-aware sample selection. By balancing successful and failed generated trajectories, PPL suppresses overly optimistic value estimates while retaining useful synthetic successes for value propagation.

\end{document}